\pdfoutput=1

\documentclass[11pt]{article}

\usepackage{acl}

\usepackage{times}
\usepackage{latexsym}

\usepackage[T1]{fontenc}

\usepackage[utf8]{inputenc}

\usepackage{microtype}

\usepackage{inconsolata}

\usepackage{amsmath}
\usepackage{amssymb}
\usepackage{mathtools}
\usepackage{amsthm}
\usepackage{bbm}
\usepackage[textsize=tiny]{todonotes}
\usepackage{tabularx}
\usepackage{booktabs}
\usepackage{multicol}
\usepackage{multirow}
\usepackage{algorithm}
\usepackage[noend]{algpseudocode}
\usepackage{comment}
\usepackage{xspace}
\usepackage{proof}
\usepackage{stmaryrd}

\usepackage{bm}
\usepackage{pifont}
\usepackage[noend]{algpseudocode}
\usepackage{wrapfig}
\usepackage{url} 
\usepackage{fixltx2e}
\usepackage{xcolor}

\newcommand\ttsmall[1]{\texttt{\small #1}}

\def\SPSB#1#2{\rlap{\textsuperscript{\textcolor{black}{#1}}}\SB{#2}}

\def\SB#1{\textsubscript{\textcolor{black}{#1}}}

\newcommand{\agre}{\textsc{Agree}}
\newcommand{\agreenotta}{\textsc{Agree}\textsubscript{w/o TTA}}
\newcommand{\agreetta}{\textsc{Agree}\textsubscript{w/ TTA}}

\newcommand{\iclcite}{\textsc{ICLCite}}
\newcommand{\postcite}{\textsc{PostCite}}
\newcommand{\postsearch}{\textsc{PostSearch}}
\newcommand{\postnli}{\textsc{PostAttr}}

\newcommand{\gscore}{\mathcal{G}}

\newcommand{\textcorpus}{\mathcal{D}}
\newcommand{\citeset}{\mathcal{C}}
\newcommand{\basellm}{\mathcal{M}^{\mathcal{B}}}
\newcommand{\adaptllm}{\mathcal{M}^{\mathcal{A}}}
\newcommand{\nlimodel}{\phi}

\algnewcommand\Input{\textbf{input: }}
\algnewcommand\Output{\textbf{output: }}

\DeclareMathOperator*{\argmax}{arg\,max}

\title{Effective Large Language Model Adaptation for Improved \\ Grounding and Citation Generation}

\author{Xi Ye$^\diamondsuit$\thanks{\,\, Work done during an internship at Google Cloud AI.} \quad Ruoxi Sun$^\spadesuit$\quad  Sercan \"{O}. Ar{\i}k$^\spadesuit$ \quad {Tomas Pfister}$^\spadesuit$  \\
$^\diamondsuit$ The University of Texas at Austin \quad $^\spadesuit$ Google Cloud AI \\
 $^\diamondsuit${\texttt{xiye@cs.utexas.edu}} \\
 $^\spadesuit${\texttt{\{ruoxis,soarik,tpfister\}@google.com}} \\
}

\begin{document}
\maketitle
\begin{abstract}

Large language models (LLMs) have achieved remarkable advancements in natural language understanding and generation. However, one major issue towards their widespread deployment in the real world is that they can generate "hallucinated" answers that are not factual.
Towards this end, this paper focuses on improving LLMs by grounding their responses in retrieved passages and by providing citations. We propose a new framework, \textit{AGREE}, \textbf{A}daptation for \textbf{GR}ounding \textbf{E}nhanc\textbf{E}ment, that improves the grounding from a holistic perspective. Our framework tunes LLMs to self-ground the claims in their responses and provide accurate citations to retrieved documents. This tuning on top of the pre-trained LLMs requires well-grounded responses (with citations) for paired queries, for which we introduce a method that can automatically construct such data from unlabeled queries.
The self-grounding capability of tuned LLMs further grants them a test-time adaptation (TTA) capability that can actively retrieve passages to support the claims that have not been grounded, which iteratively improves the responses of LLMs. Across five datasets and two LLMs, our results show that the proposed tuning-based \agre{} framework generates superior grounded responses with more accurate citations compared to prompting-based approaches and post-hoc citing-based approaches. 

\end{abstract}

\section{Introduction}

\begin{figure*}[t]
    \centering
    \includegraphics[width=0.9\linewidth, trim=0 740 75 0,clip]{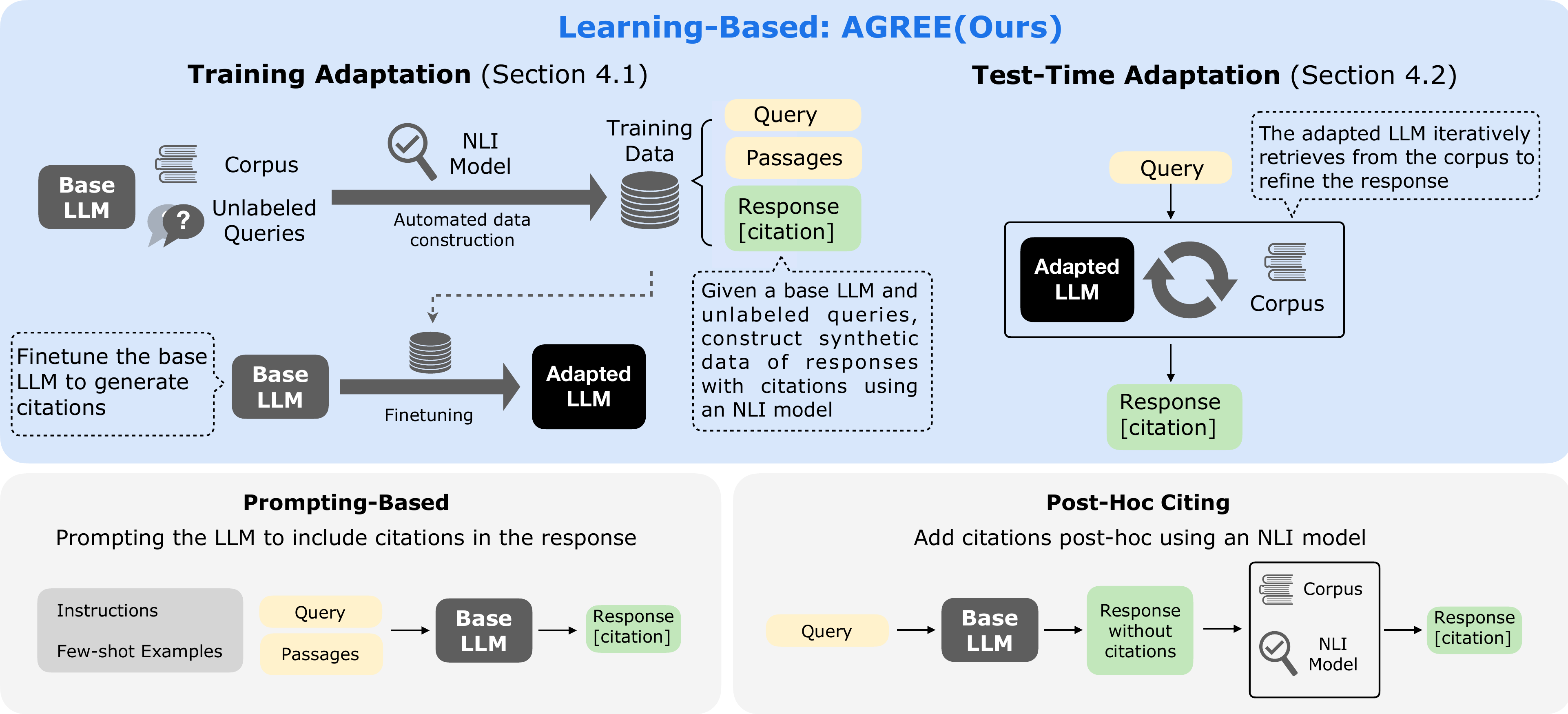}
    \caption{Our framework, \agre{}, combines tuning (Section~\ref{sec:tuning}) and test time adaptation (Section~\ref{sec:tta}) for better attribution and citation generation.}
    \label{fig:intro}
\end{figure*}

Recent advancements in large language models (LLMs) have yielded demonstrably groundbreaking capabilities in natural language processing (NLP) ~\cite{gpt3,palm}. Their ability to understand, generate, and manipulate text at unprecedented scales and depths has established them as a transformative force within the burgeoning field of artificial intelligence, poised to significantly impact our increasingly data-driven world.
Despite their widely spread adoption, one prominent issue of LLMs is that in certain scenarios they hallucinate: they generate plausible-sounding but nonfactual information~\cite{maynez-etal-2020-faithfulness,hallucinationsurvey,Menick2022TeachingLM}, limiting their the applicability in real-world settings. 
To mitigate hallucinations, solutions generally rely on grounding the claims in LLM-generated responses to supported passages by providing an attribution report~\cite{rashkin2023measuring,bohnet2022attributed,rarr} or adding citations to the claims~\cite{liuverifability,alce,Huang2023CitationAK}.

There has been a growing amount of interest in making LLM-generated responses more trustworthy by grounding and adding citations. One line of work uses instruction tuning~\cite{Kamalloo2023HAGRIDAH} or in-context learning~\cite{alce} to instruct LLMs to generate grounded responses with citations to retrieved passages, following the retrieval-augmented generation~\cite{chen2017reading,guu2020retrieval,lewis2020retrieval} framework.
As LLMs are required to perform this challenging task from just instructions and few-shot demonstrations, such directions often lead to mediocre grounding quality~\cite{alce}. Another line of work is on post-hoc citing~\cite{rarr,chen2023purr}, which links support passages to the claims in responses using a natural language inference (NLI) model. This paradigm heavily relies on LLMs' parametric knowledge and might not extend well to less-known knowledge~\cite{Sun2023HeadtoTailHK}.

We propose a new learning-based framework, \textbf{\textsc{Agree}},  \textbf{A}daptation of LLMs for \textbf{GR}ounding \textbf{E}nhanc\textbf{E}ment. As shown in Fig.~\ref{fig:intro}, our framework fine-tunes LLMs to generate citations, as opposed to prompting or relying on an external NLI model used in a post-hoc way. 
At the training phase, \agre{} collects well-grounded responses for unlabelled queries automatically from a base LLM with the help of an NLI model.
Next, the collected data are used for supervising LLMs to generate grounded responses based on the retrieved passages as well as include citations in their responses.
As a test-time approach, we propose an iterative inference strategy that allows LLMs to seek for additional information based on the self-grounding evaluation so as to refine its response.
The tuning and test-time adaptation together enable LLMs to effectively and efficiently ground their responses in the corpus.
We apply \agre{} framework to adapt an API-based LLM,
\ttsmall{text-bison}, and an open LLM, \ttsmall{llama-2-13b}, with training data collected using unlabelled queries from three datasets. 
We conduct evaluation on both in-domain and out-of-distribution datasets, comparing the proposed \agre{} framework against competitive in-context learning and post-hoc citing baselines. The experimental results highlight that \agre{} framework successfully improves grounding, in citation recall \& precision, compared to the baselines by a substantial margin (generally more than 20\%). We find LLMs can learn to add accurate citations to their responses with our carefully designed tuning mechanisms. Furthermore, the improvements in grounding quality achieved by tuning using certain datasets can generalize well across domains. To summarize, our main contributions include:

\begin{itemize}
    \item A learning-based approach that adapts a base LLM to include accurate citations in its response, leveraging automatically created data;
    \item A test-time adaptation (TTA) method that iteratively improves responses of LLMs based on the citation information;
    \item Extensive experiments on two LLMs over five datasets demonstrating the effectiveness of the proposed \agre{} framework for improving grounding and citation generation. 
\end{itemize}



\section{Related Work}
Hallucination is a prevalent issue for generative language models on many tasks~\cite{maynez-etal-2020-faithfulness,raunak-etal-2021-curious,dziri-etal-2021-neural,hallucinationsurvey,interpicl,Tang-Et-Al:2023:Factuality,Huang2023CitationAK}. It has been evaluated in different  ways, investigating the grounding in generated responses~\cite{bohnet2022attributed,rashkin2023measuring,factscore,yue2023automatic}.

Various approaches have been proposed to mitigate hallucination and improve the factuality of LLM-generated responses.
Among these, our work particularly focuses on providing citations to attributable information source~\cite{liuverifability,alce}. Unlike existing work that largely relies on zero-shot prompting or few-shot prompting~\cite{Kamalloo2023HAGRIDAH,alce} or use an additional NLI model~\cite{rarr,chen2023purr} to add citations, we propose a learning-based approach that tunes LLMs to generate better-grounded responses supported with citations.

More broadly, recent work also investigates methods for improving factuality of LLMs without using external knowledge, including inference-time intervention~\cite{li2023inferencetime,chuang2023dola}, cross-exam~\cite{Cohen2023LMVL,Du2023ImprovingFA}, self-verify~\cite{Dhuliawala2023ChainofVerificationRH}, or reinforcement learning~\cite{tian2024finetuning,wu2023finegrained}. Our work differs from them in providing citations to external knowledge in the responses. Additionally, there is past work that also uses external knowledge (e.g., knowledge base) to reduce hallucination by injecting knowledge into prompts~\cite{elaraby2023halo,peng2023check}. While the external knowledge used for generating a response can possibly serve as a coarse and general reference, these approaches also do not offer granular, sentence-level citations as in our work.

Lastly, the proposed framework is a form of a retrieval augmented generation approach. While past work has explored using retrieval to improve LLM generation quality~\cite{chen2017reading,lewis2020retrieval,guu2020retrieval,izacard2020leveraging,Shi2023REPLUGRB} or factuality~\cite{shuster-etal-2021,Jiang2023ActiveRA,panProgramFC}, our approach further enables LLMs to generate citations and self-generated citations to guide retrieval.



\section{Problem \& Background}
\label{sec:problem}
Our proposed framework aims to adapt a pre-trained LLM $\basellm$ to
$\adaptllm$ that is able to provide grounded responses with citations. Given a text query $Q$ and a corpus $\textcorpus=\{d_i\}$ consisting of text passages, the adapted LLM $\adaptllm$ is required to generate a response $A$ to the query that is factually grounded in the corpus $\textcorpus$ as well as providing citations $\citeset$ together with its response.

Following past work~\cite{liuverifability,alce}, we segment LLMs' output into statements by sentences and require each of the sentences to cite a set of passages from the corpus. Specifically, let $s_1,\ldots,s_n$ be the statements in the answer $A=s_1,\ldots,s_n$. The citations $\mathcal{C}=\{E_1,\ldots,E_n\}$ links each statement $s_i$ to a set of evidence passages $E_i\subset \textcorpus$.

Recall that our adaptation aims to provide better grounded responses. With citations $\citeset$, we can quantify the grounding quality of a response $A$ by a grounding score $\gscore$:

\small
$$\gscore(A,\citeset)=\frac{1}{n} \sum_{i} \phi(\mathrm{concat}(E_i),s_i),$$
\normalsize

\noindent where $\nlimodel$ is an NLI model that assesses whether the concatenated passage $\mathrm{concat(E_i)}$ supports $s_i$. The grounding score $\gscore$ essentially averages how well each sentence is supported by its citations.

\section{AGREE Framework}

The proposed \agre{} framework takes a holistic perspective for grounding, proposing a model tuning approach that adapts the base LLM to include citations in its responses, and introducing a test-time adaptation (TTA) mechanism that leverages the citation information for actively retrieving from the corpus and iteratively refining the responses.


\begin{figure*}[t]
    \centering

    \includegraphics[width=0.95\linewidth,trim=0 950 0 0,clip]{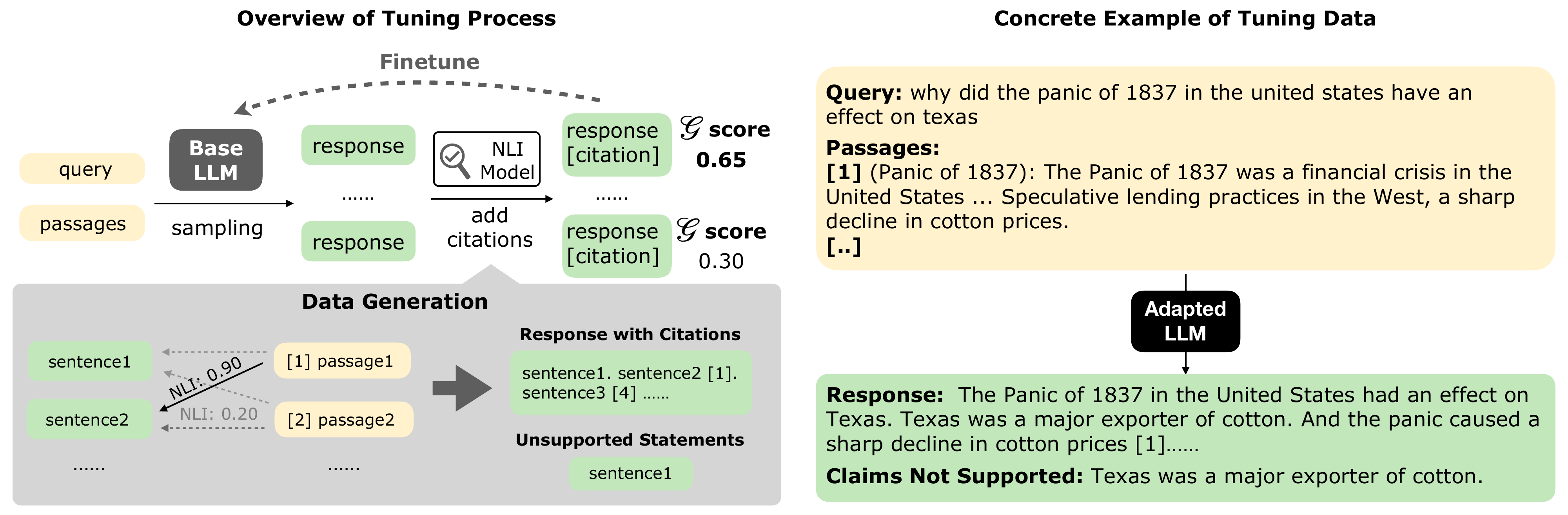}
    \caption{Illustration of the tuning process. We sample responses from the base model, use an NLI model to add citations to the sampled responses, and tune the base model with the best-grounded response. We also show a concrete example of tuning data on the right.}
    \label{fig:tuning_detail}
\end{figure*}

\subsection{Tuning LLMs}
\label{sec:tuning}


We tune the LLM to self-ground the claims in their responses by providing citations to retrieved documents.
Our method is able to grant LLMs such an ability using only a collection of \emph{unlabeled} queries $\{Q\}$ and an NLI model $\phi$. As we are using unlabeled queries without reference responses, we formulate the adaptation task as tuning LLMs to achieve better grounding without heavily deviating from the original generations (such an approach of preservation has also been adopted in recent work~\cite{rarr}). Conceptually, we adapt $\basellm$ to $\adaptllm$ so that the answers generated by the adapted LLM $\adaptllm$ should satisfy the grounding constraints (with grounding score $> \tau_{\mathcal{G}}$) while maximizing the scores with respect to the base LLM $\basellm$:

\small
\begin{equation}
\label{eq:conceptual_obj}
    \max \mathbbm{E}_{(A,\citeset) \sim \adaptllm (\cdot \mid Q,\textcorpus)} \basellm(A \mid Q,\textcorpus) \mathbbm{1}\{\gscore(A,\citeset)\ge \tau_{\mathcal{G}}\}.
\end{equation}
\normalsize

\noindent In practice, we adopt a data-centric approach for optimizing $\adaptllm$. For a given question, we opt to {\bf use the maximally-grounded response sampled from the base LLM} to construct the tuning data. We will detail the process in the following of this section.


\paragraph{Data generation}
As shown in Fig~\ref{fig:tuning_detail}, given the query, we first sample responses $\{A\}$ from the base LLM $\basellm(\cdot\mid Q,\textcorpus)$ using instruction following (see Appendix~\ref{app:detail_tuning} for details). For each $A=s_1,\ldots,s_n$ we create citations $\mathcal{C}=\{E_i\}$ using the NLI model, $\phi$, to link a sentence $s_i$ to the maximally supported passage $e_i=\max_{d_\in \textcorpus}\phi(d,s_i)$ if the passage $e_i$ actually support $s_i$ (i.e., $\phi(e_i,s_i) > \tau$).\footnote{In practice, we only present 5 passages retrieved from $\textcorpus$ to the LLM for generating initial responses, and only generate citations to this set of retrieved passages. We use TRUE~\cite{truenli}, a T5-11B NLI model. Please refer to Appendix~\ref{app:detail_tuning} for more details.} Otherwise, we do not add a citation to $s_i$, and $s_i$ is an unsupported statement. That is: $E_i = \{e_i\}\,if\, \phi(e_i,s_i) > \tau\,else\,\{\}.$ We use $U$ to denote the set of unsupported statements that cannot find citations. This allows us to evaluate the grounding of $A$ as in Section~\ref{sec:problem}. Now, we can choose the best response $A^{*}$ from $\{A\}$ based on the grounding scores to form a grounded response, i.e., $A^*=\argmax_{A} \gscore (A,C)$. 

We then use $\{Q,A^{\mathcal{*}},C^{\mathcal{*}}\}$ ($C^{\mathcal{*}}$ as the citations associated with $A^{\mathcal{*}}$) to teach the base LLM to generate grounded responses with citations. In addition to citations, we also instruct the LLM to clearly state the unsupported statements $U^*$, as shown in Fig.~\ref{fig:tuning_detail}. We note that the tuning of framework does not force all training responses to be perfectly grounded. Instead, we supervise the LLM itself to identify unsupported statements. This allows the LLM to generate more flexibly and guide the retrieval process with its knowledge.\footnote{Please refer to  Appendix~\ref{app:detail_tuning} for more details on the tuning method.}

\paragraph{Supervised fine-tuning} We have introduced how we construct supervision to instruct the LLM to add citations and state unsupported statements in its response. 
To effectively tune the LLM, we verbalize the entire process in natural language. We denote the verbalized natural language description as $\mathrm{VERB}(A^*, \citeset^*, U^*)$ (see Fig.~\ref{fig:tuning_detail} for a concrete example).\footnote{Please refer to Appendix~\ref{app:more_exs} for more examples of tuning data.} The natural language formalization also allows us to conveniently tune the LLM with standard language modeling objectives:

\small
\begin{equation}
\label{eq:actual_obj}
    \adaptllm=\argmax_{\mathcal{M}} \sum_{Q} \mathcal{M}(\mathrm{VERB}(A^*, \citeset^*, U^*) \mid Q, \textcorpus).
\end{equation}
\normalsize

\noindent We note that this actual objective, Eq~(\ref{eq:actual_obj}), maximizes the log probability of generating the best-grounded answer $A^*$ that is selected from the generation of the base model. As $A^*$ is sampled from the base model, such an objective avoids significant deviations from the original generations, which aligns with the goal of the conceptual objective (Eq~(\ref{eq:conceptual_obj})).

\paragraph{Multi-dataset training}
We use multiple existing datasets to construct the adaptation data used to tune the pre-trained LLM, including Natural Questions (NQ)~\cite{natq}, FEVER~\cite{fever}, and StrategyQA~\cite{strategyqa}. We choose these as they contain diverse text, and the answers to the corresponding queries require different types of reasoning processes: NQ provides diverse queries naturally asked by real human users; FEVER places a particular emphasis on fact verification; and StrategyQA requires multi-hop reasoning with implicit strategy. It is worthwhile to note that \agre{} \emph{only} uses queries, leaving out ground-truth answers, to improve LLMs.

\begin{algorithm*}[t]
    \scriptsize
    \caption{Iterative TTA}
    \begin{algorithmic}[1]
    \Procedure{$\textsc{IterativeInference}$}{$Q$, $\textcorpus$, $\adaptllm$, $k$, $B$}
    \Statex \Input{A query $Q$, text corpus $\textcorpus$, the adapted LLM $\adaptllm$, the number of passages $k$ that $\adaptllm$ can take as input, the budget for LLM calls $B$}
    \vspace{0.05in}
    \State $relevant\_psgs=[]$
    \State $\triangleright$ retrieve passages using the query
    \State $working\_psgs := \textsc{Retrieve}(Q, \textcorpus)[:k]$
    \State $\triangleright$ keep track of seen passages to avoid presenting duplicate passages to the LLM
    \State $seen\_psgs := []$
    \While{$iter = 1:B$}
    \State $\triangleright$ Use the LLM to generate an answer $A$ for the query $Q$ based on the working psgs $\textcorpus$. Additionally obtain the cited passages and unsupported the sentences. 
    \State $A$, $cited\_psgs$, $unsup\_sents$ $:=$ $\adaptllm$( $Q$,$working\_psgs$)
    \State $\triangleright$ add cited passages to the list of relevant passages and de-duplicate the list
    \State $relevant\_psgs := \textsc{DeDuplicate}(relevant\_psgs + cited\_psgs)$
    \State $\triangleright$ update the seen passages to include the working passages of this iteration
    \State $seen\_psgs := seen\_psgs + working\_psgs$
    \If{$unsup\_sents$ is not None}
        \State $\triangleright$ retrieve additional information related to the unsupported statements
        \State      $supplementing\_psgs := \textsc{Retrieve}(unsup\_sents, \textcorpus)$
    \Else{}
        
    \State $\triangleright$ include more query-related passages to acquire more complete information
        \State $supplementing\_psgs := \textsc{Retrieve}(Q, \textcorpus)$
    \EndIf
    \State $\triangleright$ update the working passage to include supplementing passages that have not been presented to the LLM before
    \State $working\_psgs := \textsc{DeDuplicate}(relevant\_psgs+\textsc{SetDiff}(supplementing\_psgs, seen\_psgs))[:k]$
    \EndWhile
    \State \Return $A$, $cited\_psgs$
    \EndProcedure
    \end{algorithmic}
    \label{fig:algorithm}
\end{algorithm*}

\begin{figure}[t]
    \centering
    \includegraphics[width=1.0\linewidth,trim=0 1160 1200 0,clip]{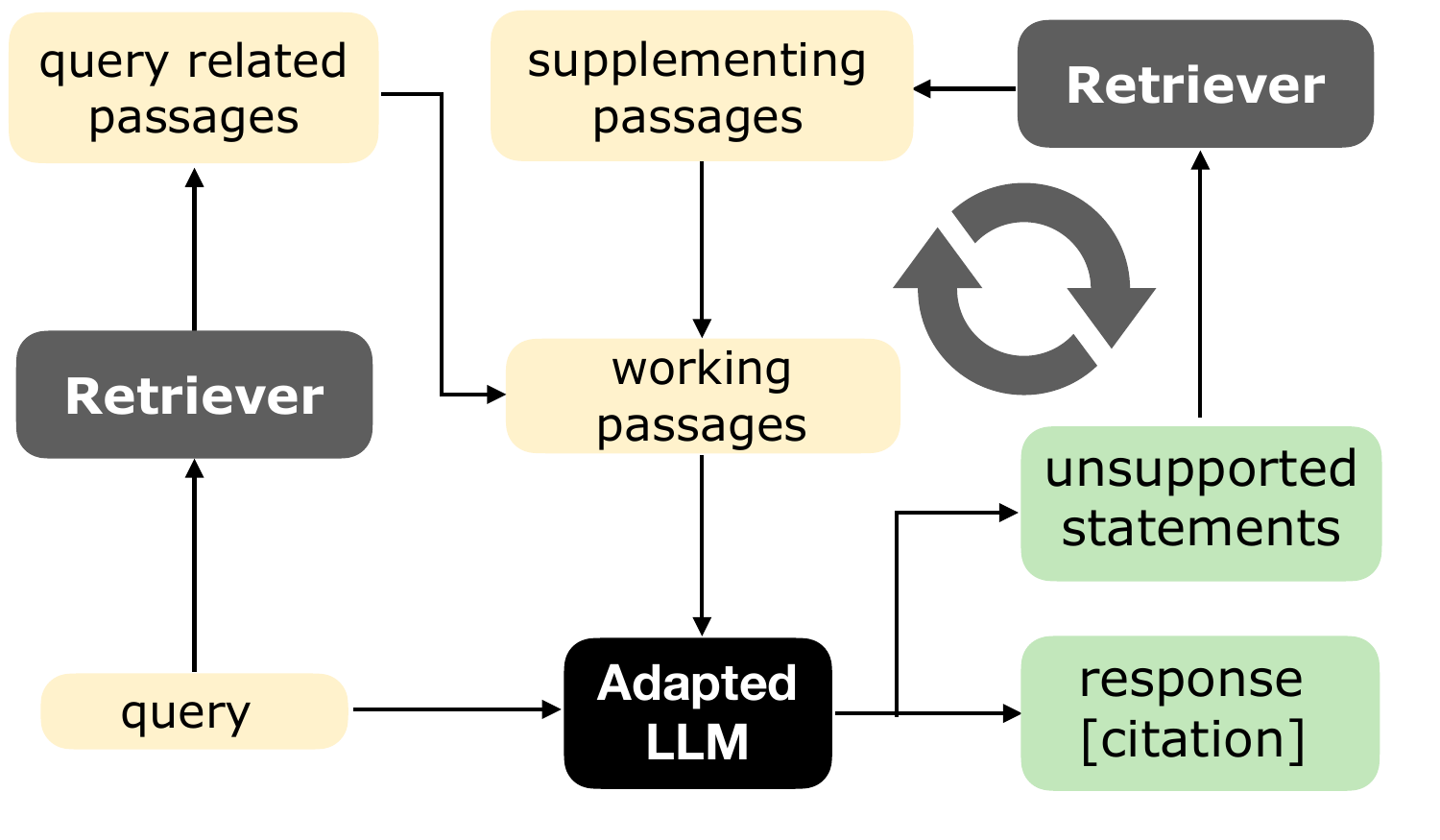}
    \caption{Illustration of the test-time adaptation mechanism. The adapted LLM retrieves from the corpus based on self-generated citation information to refine its response in an iterative way.}
    \label{fig:tta_detail}
\end{figure}

\subsection{Test-time adaptation}
\label{sec:tta}
We introduce a novel test-time adaptation (TTA) method for the inference procedure, overviewed in Fig.~\ref{fig:tta_detail}. Our framework is a form of retrieval augmented generation framework -- at the core of our approach lies the adapted LLM that is able to answer a query based on a set of given passages retrieved from the corpus, and, more importantly, self-ground its response to add citations to the passages as well as to find unsupported statements needing further investigation. With these capabilities, the adapted LLM can iteratively construct a set of relevant passages from the large corpus $\textcorpus$ and refine its response to the query.

The detailed procedure of TTA is shown in Algorithm~(\ref{fig:algorithm}). Given a query $Q$ and the corpus $\textcorpus$,
we first retrieve based on the query to obtain an initial set of working passages. Next, we employ the following procedure iteratively until we consume all the budget $B$ of invoking LLM calls. At each iteration, the LLM generates a response to the query based on the working passages, adds citations to its response, and finds out any unsupported statements that do not have citations (ln 9).
Then, we add the cited passages to the list of relevant passages. Lastly, at each iteration, we update the working passages -- if there are unsupported statements, we include additional information retrieved based on the unsupported statements (ln 15), otherwise, we include more passages that are retrieved based on the query to acquire more complete information (ln 17). We only include passages that are new and haven’t been presented to the LLM yet (ln 19). Note that at each iteration, we let the LLM to re-generate a response based on the current working passages instead of editing from previous one, which we observed lead to better fluency.

The design of our proposed TTA enables efficient and flexible inference. We rely on the LLM to generate citations itself, which has the advantage of reduced overhead of invoking an additional NLI model in a post-hoc way. Also, as we iteratively refine the answer, such a process can be streamed and flexibly controlled by setting a budget in deployment.

\section{Experiments}

\subsection{Setup}

\begin{table}[t]
    \centering
    \scriptsize
    \begin{tabular}{lccc}
    \toprule
    Dataset & Type & Corpus & \# \\
    
    \midrule
    & \multicolumn{3}{c}{Train} \\
         NQ & Factoid QA &  Wiki & 2500 \\
         StrategyQA& Multi-htop QA & Wiki & 1000\\
         Fever & Fact Checking & Wiki & 1000 \\
    \cmidrule{2-4}
   & \multicolumn{3}{c}{In-Distribution Test} \\
    NQ & Factoid QA &   Wiki & 700 \\
    StrategyQA & Factoid QA &  Wiki & 460 \\
    \cmidrule{2-4}
       & \multicolumn{3}{c}{Out-of-Distribution Test} \\
    ASQA & Ambiguous QA &  Wiki &948 \\ 
    QAMPARI & Multi-answer QA & Wiki & 1000 \\
    Enterprise & Customer Support QA &  Enterprise &580 \\
    \bottomrule
        
    \end{tabular}
    \caption{Statistics used for adaptation and test datasets. In addition to in-domain test datasets, we also investigate the generalization to out-of-distribution datasets that exhibit different reasoning processes or different corpus types.}
    \label{tab:dataset_stats}
\end{table}

\begin{table*}[t]
    \centering
    \footnotesize
      \renewcommand{\tabcolsep}{1.10mm}
    \begin{tabular}{lcccccccccccccccccccc}
    \toprule
        &&& \multicolumn{3}{c}{NQ} && \multicolumn{3}{c}{StrategyQA} && \multicolumn{3}{c}{ASQA} && \multicolumn{3}{c}{QAMPARI} && \multicolumn{2}{c}{Enterprise} \\
         & && em-rec & rec & pre && acc & rec & pre && em-rec & rec & pre && rec-5 & rec & pre && rec & pre \\

         \cmidrule{4-6} \cmidrule{8-10} \cmidrule{12-14} \cmidrule{16-18} \cmidrule{20-21}
        &&&& \multicolumn{16}{c}{Base model: \texttt{text-bison-001}}\\

        \iclcite{} & &&  47.6 &	52.1 & 56.3	&&	74.5 & 13.6	& 27.8	&&	39.5 & 47.3	 & 49.8	&&	20.3 & 22.7&	24.5	&&	30.2 & 40.5 \\
        \cmidrule{1-2}
        \postsearch{} & && 45.1 & 29.7 &	28.7	&& \bf	75.5 &	20.1 &	20.1 &&	38.4 & 19.2	 & 19.2	&&	\bf 22.5 & 16.2	& 16.2	&&	15.9	& 15.9\\
        \postnli{} & && 45.1 &31.5 &	31.5&&	75.5 & 18.4 &	18.4	&&	35.1 & 38.0 & 38.0	&&	22.5 & 18.5 &	18.5	&&	20.1 & 20.1 \\
        \cmidrule{1-2}
        \agre{}\SB{\sc w/o TTA} & && 50.0 &	67.9 &	73.1	&&	74.1 &	33.4	& 50.5	&&	39.5 & 65.9	& 70.5	&&	20.1 &	60.1 & 64.5	&&	55.8 & 67.1\\
        \agre{}\SB{\sc w/ TTA}& && \bf 53.1	& \bf 70.1& \bf 75.0	&&	74.9 & \bf	39.2 & \bf	57.9	&&	\bf 40.9&\bf 73.2& \bf 77.0	&&	20.9& \bf 62.9&	\bf 67.1&&	\bf 57.2& \bf 68.6 \\

                 \cmidrule{4-6} \cmidrule{8-10} \cmidrule{12-14} \cmidrule{16-18} \cmidrule{20-21}  

                &&&& \multicolumn{16}{c}{Base model: \texttt{llama-2-13b}}\\
        \iclcite{} &  &&  45.8	& 42.8 &	41.6	&&	\bf 65.5 &	20.6 &	33.1	&&	35.2 &	38.2&	39.4	&&	\bf 21.0 &	10.2 &	10.4	&&	30.6& 	38.8\\
        \cmidrule{1-2}
        \postsearch{} & && 35.9	&17.5	&17.5	&&	64.3&	8.7&	8.7	&&	25.0 &	23.6 &	23.6	&&	12.0 &	27.5 &	27.5	 &&	13.4	& 13.4 \\
        \postnli{} & && 35.9	&26.0	& 26.0	&&	64.3&	12.5&	12.5	&&	25.0 &	33.6 &	33.6	&&	12.0 &	28.9 &	28.9	 &&	18.7	& 18.7 \\
        \cmidrule{1-2}
        \agre{}\SB{\sc w/o TTA} &  && 47.9	&  50.5 &  56.6	&&	65.0 &25.5 &35.0	&&	35.7  &	 50.2&	55.3	&&	17.1& 	40.4	& 43.6	&&	 \bf 50.6&	53.8\\
        \agre{}\SB{\sc w/ TTA} & && \bf  51.0 &	\bf 62.0 & 	\bf 66.0	&&	64.6 &	\bf   30.2 & 	\bf  37.2	&&	\bf 39.4 & \bf 64.0 & \bf  66.8	&&	17.9  & \bf 51.4 & \bf 53.4	&&	50.4 &  \bf 55.4 \\
        \bottomrule
    \end{tabular}
    \caption{Answer accuracy and grounding (measured by citation quality) of \agre{} and baselines across 5 datasets. Our approach achieves substantially better citation grounding (measured by citation recall) and citation precision compared to the baselines.}
    \label{tab:main_citation}
\end{table*}

\paragraph{Evaluation datasets}
We conduct comprehensive evaluation on 5 datasets. Recall that we train \agre{} on multiple datasets including NQ, StrategyQA, and Fever. In addition to the two in-domain test sets, NQ and StrategyQA (we leave out the non-QA dataset, FEVER), we further test the generalization of adapted LLMs on 3 out-of-domain datasets, including ASQA~\cite{asqa}, QAMPARI~\cite{qampari}, and an Enterprise dataset.\footnote{We use FEVER to create tuning data, but do not use it for evaluations. As we use LLMs in a zero-shot setting, the LLMs do not always answer with the specific labels defined in FEVER, which might introduce inaccuracies in the evaluation of answer correctness.} In particular, ASQA and QAMPARI contain questions of ambiguous answers and multiple answers. The Enterprise dataset is a proprietary dataset which requires provided answers that are grounded in customer service passages. Such an evaluation suite allows assessing the generalization capability of the adapted LLMs for OOD question types (ASQA and QAMPARI) as well as to an entirely different corpus (Enterprise).

\paragraph{Models}
We demonstrate \agre{} framework with two LLMs, \ttsmall{text-bison} and \ttsmall{LlaMA-2-13B}~\cite{touvron2023llama}. We use GTR-large~\cite{gtr} as our retriever, and use TRUE~\cite{truenli} as the NLI model.

\paragraph{Baselines}
We evaluate the effectiveness \agre{} in two settings, invoking LLMs once, without TTA; and invoking LLMs multiple times, with the proposed TTA.\footnote{We set the budget $B$ for LLM calls used in TTA to be 4.} We compare with three baselines from recent work, including one prompting-based approach and two post-hoc citing approaches, described below.

\textbf{Few-shot In-Context Learning} (\iclcite{}): Following~\citet{alce}, we prompt LLMs with few-shot examples~\cite{alce}, each consisting of a query, a set of retrieved passages, and an answer with inline citations. The LLMs can therefore learn from the in-context examples and generated citations in the responses. It is worthwhile to note that \emph{ \iclcite{} is a RAG baseline that also uses retrieved passages}.

\textbf{Post-hoc search} (\postsearch{}): Following~\citet{alce}, given a query, we first instruct LLMs to answer the query \emph{without} passages, and then add citations in a post-hoc way via searching.  We link each claim in the response to the most relevant passage retrieved from a set of query-related passages.
This baseline only uses the retriever but not the NLI model.

\textbf{Post-hoc Attribution} (\postnli{}): Following~\citet{rarr}, instead of citing the most relevant passage, for each claim, we retrieve a set of $k$ passages from the corpus, and then use the NLI model, $\nlimodel$, to link to the passage that maximally supports the claim. We note both baselines in the post-hoc citing paradigm only rely on LLMs' parametric knowledge.\footnote{Please refer to Appendix~\ref{app:detail_exp} for more details on experimental setup.}




\paragraph{Metrics} We mainly focus on improving the grounding quality of generated responses, reflected by the quality of citations. Following past work~\cite{alce}, we report the \textbf{citation recall} (rec) and \textbf{citation precision} (pre) on all the evaluation datasets. We note that \textbf{citation recall aggregates how well each sentence is supported by the citation to the corpus, which is essentially the grounding score $\gscore$}. Therefore, we prioritize on the evaluation of citation recall.

We also report the correctness of the generated outputs. For NQ, we report exact match recall (em-rec; whether the short answers are substrings in the response). For StrategyQA, we report the accuracy (acc). For ASQA and QAMPARI, we use subsets from~\citet{alce}, and report the exact match recall (em-rec) for ASQA and recall-5 (rec-5, considering recall to be 100\% if the prediction includes at least 5 correct answers) for QAMPARI. For the Enterprise dataset, we only report the citation quality as there are no ground truth answers for this dataset, and citation quality reflects whether the model can provide accurate information.

\subsection{Results and analyses}

\paragraph{Tuning is effective for superior grounding:} Table~\ref{tab:main_citation} summarizes the results obtained using our \agre{} framework and compares with the baselines. As suggested by the results, across 5 datasets, \agre{} can generate responses that are better grounded in the text corpus and provide accurate citations to its response, substantially outperforming all the baselines. When tuned with high-quality data, LLMs can effectively learn to self-ground their response without needing an additional NLI model. On the other hand, \iclcite{}, which solely relies on in-context learning, cannot generate citations as accurately as a tuned LLM, as suggested by the large gap on citation precision between \iclcite{} and \agre{}. We also observe similar findings as suggested by \citet{alce}: \postcite{} often leads to poor citation quality -- without being conditioned on passages, the response from \postcite{} often cannot be paired with passages that lead to high citation recall for the generated claims.


\begin{table*}[t]
    \centering
    \scriptsize
      \renewcommand{\tabcolsep}{1.25mm}
    \begin{tabular}{lcccccccccccccccccccc}
    \toprule
        &&& \multicolumn{3}{c}{NQ} && \multicolumn{3}{c}{StrategyQA} && \multicolumn{3}{c}{ASQA} && \multicolumn{3}{c}{QAMPARI} && \multicolumn{2}{c}{Enterprise} \\

         &  && em-rec & rec & pre && acc & rec & pre && em-rec & rec & pre && rec-5 & rec & pre && rec & pre \\
        \cmidrule{1-2} \cmidrule{4-6} \cmidrule{8-10} \cmidrule{12-14} \cmidrule{16-18} \cmidrule{20-21}  
        &&&& \multicolumn{16}{c}{Base model: \texttt{text-bison-001}}\vspace{0.25em}\\

        \iclcite{} & &&  47.6 &	52.1 & 56.3	&&	\bf 74.5 & 13.6	& 27.8	&&	39.5 & 47.3	 & 49.8	&&	20.3 & 22.7&	24.5	&&	30.2 & 40.5 \\
        \cmidrule{1-2}
        \agre{}\SPSB{\bf Multi-dataset}{\sc w/o TTA} & && \bf 50.0 & \bf 67.9 & \bf	73.1	&&	74.1 &	\bf 33.4	& \bf 50.5	&& \bf	39.5 & \bf 65.9	& \bf 70.5	&& \bf	20.1 &	\bf 60.1 & \bf 64.5	&&	\bf 55.8 & \bf 67.1\\
       \agre{}\SPSB{\bf NQ-only}{\sc w/o TTA}& &&  49.4	& 62.3 & 69.1	&&	74.1&33.0 & 45.5	&&	38.4 &56.0&64.5	&&	19.1&43.7&49.5	&&	40.5&59.2 \\
        \cmidrule{1-2} \cmidrule{4-6} \cmidrule{8-10} \cmidrule{12-14} \cmidrule{16-18} \cmidrule{20-21}  
        &&&& \multicolumn{16}{c}{Base model: \texttt{llama-2-13b}}\vspace{0.25em}\\
        
         \iclcite{} &  &&  45.8	& 42.8 &	41.6	&&	 65.5 &	20.6 &	33.1	&&	35.2 &	38.2&	39.4	&& 21.0 &	10.2 &	10.4	&&	30.6& 	38.8\\
        \cmidrule{1-2}
        \cmidrule{1-2}
       \agre{}\SPSB{\bf Multi-dataset}{\sc w/o TTA}&  && 47.9	&  50.5 &  56.6	&&	65.0 &25.5 &35.0	&&	 35.7 &	 50.2&	 55.3	&&	17.1& 	40.4	& 43.6	&&	 50.6&	52.8\\
      \agre{}\SPSB{\bf NQ-only}{\sc w/o TTA}& &&  48.1 & 47.4 & 53.6 &&	62.1 & 25.0 & 30.2	&&	35.0 & 44.0 & 51.2	&&	15.7 & 33.1 & 38.0 &&	44.7 & 49.2\\
             \cmidrule{1-2}
      \agre{}\SPSB{\bf Distill}{\sc w/o TTA} & && 47.9	& 59.1	& 65.1	&&	64.4 & 30.5 & 41.1 &&	35.2 &58.5 &65.2	&&	17.9 & 52.5	& 52.7	&&	48.1 & 55.9 \\
        
        \bottomrule
    \end{tabular}
    \caption{Analysis on the impact of training data. Training with multiple datasets ({\agre{}\textsuperscript{Multi-dataset}}) leads to better grounding (citation recall) and better citation precision across datasets, compared to training using the NQ dataset (\agre{}\textsuperscript{NQ-only}). The citation quality of a less capable model \ttsmall{llama-2-13b} can also benefit from tuning using outputs from a more capable model (\ttsmall{text-bison-001}).}
    \label{tab:analysis_training}
\end{table*}

\begin{table}[t]
    \centering
    \scriptsize
    \begin{tabular}{lcc}
    \toprule
     & \# Tok: LLM &\# Tok: NLI (T5-11B) \\
     \midrule
        \iclcite{} & 2800 & $-$\\
        \postnli{} & 360 & 3520\\
                \cmidrule{1-1}
       \agreenotta{} & 1210 & $-$\\
        \agreetta{} & 4840& $-$\\
         \bottomrule
    \end{tabular}
    \caption{The average computation cost (for one query) of different methods measured by the number of tokens processed by the LLM and the NLI model (based on a T5-11B architecture). \agreenotta{} is able to achieve better citation quality compared to \iclcite{}, despite consuming less than half of the tokens needed for \iclcite{}.}
    \label{tab:cost_of_toks}
\end{table}

\paragraph{The performance improvements can generalize:} Recall that we adapt the base LLM only using in-domain training sets (NQ, StrategyQA, and FEVER), and directly test the model on out-of-distribution (OOD) test set (ASQA, QAMPARI, Enterprise). The results suggest that the improvements obtained from training on in-domain datasets can effectively generalize to OOD datasets that contain different question types or use different types of corpus. This is a fundamental advantage of the proposed approach -- \agre{} can generalize to a target domain in the zero-shot setting without needing any samples from the target domain, which is needed for \iclcite{}.

\paragraph{TTA improves both grounding and answer correctness:} The comparison between \agre{} without and with TTA highlights the effectiveness of the proposed iterative TTA strategy. We observe improvements in terms of both better grounding and accuracy. For instance, TTA improves \ttsmall{llama-2} answer correctness by 3.1 and 3.7 on NQ and ASQA, respectively. Such improvements can be attributed to the fact that our TTA allows the LLMs to actively collect relevant passages to construct better answers following the self-grounding guidance.

\paragraph{Discussions on answer correctness:} In general, \agreetta{} can achieve better correctness compared to \iclcite{}. \agreenotta{} achieves similar answer correctness with \iclcite{}, as both methods are conditioned on the same set of passages. As a result, the quality of passages heavily intervenes on the correctness of the answers.
Unlike \agre{} and \iclcite{}, \postnli{} purely relies on the parametric knowledge of the LLMs to answer the query.
As a result, \postnli{} generally achieves inferior answer correctness compared to \agre{} and \iclcite{} on these two LLMs, especially on the less capable LLM, \ttsmall{llama-2-13b}, that has less accurate knowledge compared to \ttsmall{bison}. Moreover, on the Enterprise dataset which contains more domain-specific information, \postnli{} utterly fails to recall attributable information from LLMs' parametric knowledge.


\paragraph{Results with different LLMs:} 


Our approach successfully adapts both \ttsmall{text-bison-001} and \ttsmall{llama-2-13b}.
\ttsmall{llama} is generally less capable compared to \ttsmall{bison}, underperforming  \ttsmall{bison} in terms of answer correctness and citation quality. Still, \agre{} also consistently outperforms the baseline, generating more grounded answers as well as providing more precise citations. This highlights that the proposed tuning-based adaptation approach is model-agnostic and is effective across LLMs of varying capabilities.


\paragraph{Computational efficiency:} \agre{} framework fine-tunes the base LLM to enable self-grounding without needing for additional in-context examples or NLI models. As a result, our framework is able to achieve strong citation performance without expensive inference cost.
Table~\ref{tab:cost_of_toks} shows the comparison between the computation cost, measured by the number of tokens processed by the LLM and the NLI model, needed for one query of our methods and that of the baselines. Compared to \iclcite{}, \agreenotta{} uses much fewer tokens due to not using additional in-context examples, but achieves significantly better citation quality (see Table~\ref{tab:main_citation}). \postnli{} does not use retrieved passages in the prompts and hence requires less computation on the LLM compared to our framework, but it requires additional overhead of extensively invoking the NLI model (which has 11B parameters -- see Appendix~\ref{app:detail_tuning} for details) to verify the each of the claims based on each of the retrieved passages. The citation performance of \postnli{} also substantially lags \iclcite{} and \agre{}. \agreetta{} requires more computation compared to \agreenotta{}, but is able to achieve both better citation quality and improvements in answer correctness.

\paragraph{The impact of training with multiple datasets:} \agre{} uses multiple datasets spanning factoid QA, multi-hop reasoning, and fact-checking to construct data for adapting the base model. We expect such a combination can grant the adapted model better generalization to different types of questions and different text distributions. We conduct an analysis to investigate the benefits of using multiple datasets for tuning. Table~\ref{tab:analysis_training} shows the performance of our approach trained using multi-datasets and a counterpart that is trained only on NQ data (\agre{}\textsuperscript{NQ-only}). The results suggest that training using NQ leads to inferior citation quality compared to training on the combination of three datasets across all the datasets. The performance gap is especially significant on datasets other than NQ. Moreover, training only on NQ also leads to inferior answer correctness across all the datasets. Nevertheless, training only on NQ can still improve performance compared to solely relying on in-context learning (\iclcite{}).

\paragraph{\agre{} as a distillation approach:} Our work mainly focuses on improving the base LLM's grounding capability in a self-improving way. That is, we use the samples generated by the base LLM itself to adapt the base LLM, as opposed to distilling from proprietary models which may impose constraints on the deployment of adapted model. 
Nevertheless, we conduct an analysis to investigate the effectiveness as a distillation approach in a data-centric way by using data generated from a more capable model to enhance the grounding data for a less capable model. We use the data generated by \ttsmall{text-bison-001} to tune \ttsmall{llama-2-13b}.\footnote{Please refer to Appendix~\ref{app:detail_exp} for details.}  As shown in the last row of Table~\ref{tab:analysis_training}, \agre{}\textsuperscript{Distill} achieves better citation quality compared to \agre{} on \ttsmall{llama-2-13b}, as it is trained on better grounded responses produced by a more capable model. However, \ttsmall{llama-2} fine-tuned on data generated by \ttsmall{bison} still can't level the performance on \ttsmall{bison} as constrained by the gap between the capabilities of these two LLMs.

\paragraph{Qualitative analyses:}

\begin{figure}[t]
    \scriptsize
    \begin{tabular}{|l|}
    \hline
    \multicolumn{1}{|c|}{\bf Example from NQ} \\
    \hline
    \textbf{Query:}  where did the world`s largest recorded wave occur? \vspace{0.25em} \\ 
     \multicolumn{1}{|c|}{\bf Without TTA} \\
    \textbf{Passage [1]:} \textcolor{gray}{(retrieved using the query)} \\
    (Title: 1958 Lituya Bay earthquake and megatsunami) Lituya Bay has a \\ history of megatsunami events, but the 1958 event was the first for which\\ sufficient data was captured... \\
    \textbf{\agreenotta{}:} \underline{The world's largest recorded wave occurred in Lituya} \\
    \underline{Bay,Alaska.} The data was captured in a 1958 megatsunami event [1]. \vspace{0.25em} \\

        \multicolumn{1}{|c|}{\bf With TTA} \\

    \textbf{Passage [2]:} \textcolor{gray}{(retrieved using  \underline{prior unsupported statement})} \\
    (Title: Megatsunami) In 1958, a 7.8 strike-slip earthquake in southeast \\ Alaska caused 90 million tonnes of rock and ice to drop into the deep\\ water at the head of Lituya Bay.
    The block fell almost vertically and hit \\ the water
    with sufficient force to ... \\
    \textbf{Passage [3]:} \textcolor{gray}{(retrieved using  \underline{prior unsupported statement})} \\
    (Title: Tsunami) ...Their existence was confirmed in 1958, when a giant \\ land
    slide in Lituya Bay, Alaska, caused the highest wave ever recorded, \\ which had
    a height of 524 metres.... \\
    \textbf{\agreetta{}:} The world's largest recorded wave occurred in Lituya \\ Bay, Alaska [3]. The wave was caused by a giant landslide that occurred \\ in 1958 [2]. The landslide displaced 90 million...[2].  \vspace{0.25em} \\
    \hline
    \multicolumn{1}{|c|}{\bf Example from ASQA} \\
    \hline
    \textbf{Query:} Who played the weasley brothers in harry potter?  \vspace{0.25em} \\ 
     \multicolumn{1}{|c|}{\bf Without TTA} \\
        \textbf{Passage [1]:} \textcolor{gray}{(retrieved using the query)} \\
        (Title: James and Oliver Phelps) James Andrew Eric and Oliver Martyn \\John Phelps are identical twin British actors, best known for playing \\Fred and George Weasley in the "Harry Potter" film series.\\
 \textbf{\agreenotta{}:} The Weasley brothers in Harry Potter were played by\\ identical twin actors James and Oliver Phelps [1]. \vspace{0.25em} \\
  \multicolumn{1}{|c|}{\bf With TTA} \\
         \textbf{Passage [2]:} \textcolor{gray}{(additional passages retrieved using the query)} \\
         (Title: Chris Rankin) ... Chris Rankin Christopher William "Chris" \\ Rankin is a New Zealand-born British actor who is best known for playing \\Percy Weasley in the "Harry Potter" film franchise.\\
        \textbf{\agreetta{}:} James and Oliver Phelps played Fred and George Weasley \\ in the Harry Potter film series [1]. Chris Rankin played Percy Weasley in \\ the Harry Potter film franchise [2].  \vspace{0.25em} \\
    \hline
    \end{tabular}
    \caption{Output examples of the proposed \agre{} framework with \ttsmall{text-bison-001} as the base model. TTA is able to improve the response by retrieving more relevant information to precisely support a statement (see top) or finding more passages to generate a more complete response (see bottom).}
    \label{fig:quali_exs}
\end{figure}

We qualitatively analyze the advantages of the proposed \agre{} framework compared to \iclcite{}, the strongest among the baselines. We observe that on both \ttsmall{text-bison-001} and \ttsmall{llama-2-13b}, \iclcite{} achieves inferior citation quality due to failure in following the citation format (e.g., adding citations after the periods, violating the instructions), linking a statement to a relevant but un-attributable passage (as indicated by poor citation precision), and introducing more auxiliary information not mentioned in the retrieved passages (as indicated by citation recall). Our \agre{} framework mitigates these issues by tuning on well-grounded responses certified by the NLI model. We also provide example outputs in Fig.~\ref{fig:quali_exs} comparing the outputs of \agre{} with and without proposed TTA and observe that TTA can help find more supporting passages by active retrieving using unsupported statements (top) or iteratively find more passages to construct a more complete response (bottom). 

\section{Conclusion}
We introduce a novel framework, \agre{}, that adapts LLM for improved grounding. The proposed framework tunes a pre-trained LLM to self-ground its response in retrieved passages using automatically collected data. The integrated capability for grounding their responses further enables the LLM to improve the responses at test time. Our evaluations across five datasets demonstrate the benefits of the proposed learning-based approach compared to approaches that solely rely on prompting or the parametric knowledge of LLMs.

\section{Limitations and future work}
\agre{} employs an automated data creation that relies on an NLI model, instead of humans. Thus, the citation quality is dependent on the NLI model. As suggested in \citet{alce,truenli}, one issue might be favoring ``fully support'' and cannot effectively detect ``partially support". Thus, the adapted LLMs may favor adding ``fully support" citations. One solution is to curate a set of human-annotated citations for ``partially support", which we defer to future work.
Also, our evaluation follows prior work~\cite{rashkin2023measuring,rarr} and uses the NLI model to quantify the grounding and citation quality. Therefore, our work can encounter the same issue as past work: grounding and citation quality evaluation is limited by the capabilities of the NLI model.

\agre{} uses created grounded responses to LLMs via supervised fine-tuning, as we demonstrate it leads to strong empirical results. It is also possible to treat grounding as a preference and RLHF~\cite{Ouyang2022TrainingLM} to tune LLMs, which we leave to future work.
\agre{} tuning incurs additional cost that is a one-time requirement for adapting the LLM. Considering the substantial grounding improvements, we believe this would be acceptable for most applications, especially for those with high-reliability requirements. Future work can possibly explore training a separate universal improved grounding model beyond task-specific adaptation.

We have mainly considered open domain question answering datasets focusing on information seeking tasks in English. Generalization to other long form generation tasks and other languages can be important future work directions.

Lastly, adding citations to LLM-generated responses in \agre{} might carry a shared risk with related research -- a seemingly plausible but incorrect citation could  potentially make an unsupported statement more convincing to users. 

\section*{Acknowledgments}

 Thanks to anonymous reviewers for their helpful feedback, as well as to Jinsung Yoon, Andreas Terzis, Yanfei Chen, Ankur Taly, Lucas Zhang, and Tina Pang for their help with various aspects of this work.

\bibliography{custom}

\appendix
\onecolumn
\newpage
\twocolumn

\section{Details of Data Generation}
\label{app:detail_tuning}

Recall that we create the tuning data by first sampling responses from the base LLM and then using the NLI model to create citations and identify unsupported statements. We provide the details on the process in the following of this section.

\paragraph{NLI model} We use TRUE NLI~\cite{truenli} as, to the best of our knowledge, it is the state-of-the-art NLI model for evaluating whether a passage supports a claim, and it is commonly used in the recent line of work on attributed QA and evaluating grounding~\cite{bohnet2022attributed,he2022rethinking,halueval}.\footnote{ \url{https://huggingface.co/google/t5_xxl_true_nli_mixture}} TRUE is trained on data collected from 6 datasets of diverse tasks covering NLI, paraphrase detection, and fact verification, which leads to its strong performance across diverse types of text. Furthermore, the citation performance evaluated by TRUE highly aligns with human evaluation~\cite{alce}.

\paragraph{Corpus \& retriever} As mentioned before, our framework is an instantiation of retrieval-augmented framework. For the datasets using Wikipedia as the corpus (NQ, StrategyQA, ASQA, and Qampari), we use the 2018-12-20 Wikipedia snapshot as the corpus and set up the retriever using GTR-large~\cite{gtr}.

\begin{figure}[h]
    \centering
    \small
    \begin{tabularx}{\linewidth}{|X|}
         \toprule
         Task: You will be given a question and some search results. Please answer the question in 3-5 sentences, and make sure you mention relevant details in the search results. You may use the same words as the search results when appropriate. Note that some of the search results may not be relevant, so you are not required to use all the search results, but only relevant ones. \\
\\
<Question> \\
 \\
Search Results: \\
{[}<Index>{]} <Title> \\
<Text>\\
\\
{[}...{]}\\
\\
Answer:
\\
         \bottomrule
    \end{tabularx}
    \caption{Zero-shot prompt template for sampling initial responses from the base LLM.}
    \label{fig:init_sample_prompt}
\end{figure}

\paragraph{Sampling initial responses} We sample initial responses from the base LLM using instruction following in a \emph{zero-shot fashion}. Given a query, we present the base LLM with query and 5 retrieved passages appended after an instruction that requires the base LLM to answer the query based on the passages; see Fig.~\ref{fig:init_sample_prompt} for the template of the zero-shot prompt. We note that we opt to use a zero-shot prompt as opposed to a task-specific few-shot prompt since 1) this can avoid biasing the generation with the few-shot in-context examples, and 2) this matches the expected scenario for deploying the adapted LLM to handle new queries in a zero-shot fashion.

For \ttsmall{text-bison-001}, we sample 4 responses using a temperature of 0.5. For \ttsmall{llama-2-13b}, we sample 4 responses using nuclear sampling~\cite{holtzman2019curious} with p=0.95.

\paragraph{Adding citations and identifying unsupported statements} After obtaining the initial response $\{A\}$ from the base LLM. We break each response $A$ into sentences into $s_1,\ldots,s_i$. For each $s_i$, we find the maximally supported passage $e_i$ (scored by $\phi(e_i,s_i)$) that the base LLM has seen during generating the initial responses. We link $e_i$ to $s_i$ if $\phi(e_i,s_i) > 0.7$ to encourage more precise citations. For a sentence $s_i$ if there does \emph{not} exist an $e_i$ such that $\phi(e_i,s_i) > 0.5$ (the decision boundary for entailment), we add $s_i$ to the unsupported statement set $U$.

\begin{figure}[t]
    \centering
    \small
    \begin{tabularx}{\linewidth}{|X|}
         \toprule
         \multicolumn{1}{|c|}{\bf Input} \\
        \midrule
         Task: You will be given a question and some search results. You are required to perform the following steps.\\
         \\
         First, please answer the question in 3-5 sentences, and make sure you mention relevant details in the search results. You may use the same words as the search results when appropriate. Note that some of the search results may not be relevant, so you are not required to use all the search results, but only relevant ones.  If you use the provided search results in your answer, add [n]-style citations. \\
        \\
        Next, review your response and find the unsupported sentences that do not have citations.\\
\\
<Question> \\
 \\
Search Results: \\
{[}<Index>{]} <Title> \\
<Text>\\
\\
{[}...{]}\\
\midrule
          \multicolumn{1}{|c|}{Output} \\
          \midrule
          Answer: <Response with citations>\\
          \\
          Sentences Not Supported by Citations: <Unsupported statements> \\
         \bottomrule
    \end{tabularx}
    \caption{Verbalization template for creating the training data for adapting the base LLM.}
    \label{fig:tuning_prompt}
\end{figure}

\paragraph{Verbalizing}
We show the template for verbalizing the data used to tune the LLM in Fig.~\ref{fig:tuning_prompt}. As shown in the figure, we verbalize the citations in enclosed box brackets that are added at the end of sentences (before periods) like [n], and verbalize unsupported statements after the responses.

\section{Details of Experimental Setup}
\label{app:detail_exp}

\paragraph{Details of Finetuning} For tuning, we use LORA tuning~\cite{hu2022lora} in experiments on both \ttsmall{text-bison-001} and \ttsmall{llama-2-13b}. For \ttsmall{bison}, we use API to perform tuning.\footnote{https://cloud.google.com/vertex-ai/docs/generative-ai/models/tune-text-models-supervised} and follow all the default hyper-parameters except for training steps. We set 10\% data created as development data and choose to use a training step of 1000 (chosen from 500, 1000, and 2000). For \ttsmall{llama-2}, we use the huggingface transformers~\cite{wolf2019huggingface} chat-version checkpoint.\footnote{https://huggingface.co/meta-llama/Llama-2-13b-chat-hf} We find the chat-version achieves better performance than the base checkpoint in our preliminary investigation. We set lora\_r to be 32, and only choose to use a learning rate of 1e-5 (chosen from 1e-4 and 1e-5) using the development set. We fine-tune \ttsmall{llama-2} on two A100 (40GB) GPU for 4 epochs.

\paragraph{Details of Decoding} Our evaluation uses the official implementation from ALCE~\cite{alce}, we use the same data split and prompt template from ALCE. We use temperature 0.25 for evaluation on both \ttsmall{bison} and \ttsmall{llama}. We use one sample for evaluation since adapted LLMs tend to generate better-grounded response exhibiting less variation.

\paragraph{Details of Distillation} For distillation, we directly tune \ttsmall{llama-2-13b} using the data created with \ttsmall{text-bison-001}. We also set lora\_r to be 32, use a learning rate of 1e-5, and fine-tune for 4 epochs.

\section{Comparison to \iclcite{} on More Capable LLMs}


\begin{table}[h]
    \centering
    \footnotesize
      \renewcommand{\tabcolsep}{1.10mm}
    \begin{tabular}{lccccccc}
    \toprule
        & \multicolumn{3}{c}{ASQA} && \multicolumn{3}{c}{QAMPARI} \\
        \cmidrule{2-4}\cmidrule{6-8}
       &  em-rec & rec & pre && rec-5 & rec & pre \\
                & \multicolumn{7}{c}{Base model: \texttt{llama-2-13b}}\vspace{0.25em}\\


        \agre{}\SB{\sc w/o TTA} &35.7  &	 50.2&	55.3	&&	17.1& 	40.4	& 43.6	\\
        \agre{}\SB{\sc w/ TTA} 	& 39.4 &  64.0 &  66.8	&&	17.9  & \bf 51.4 & \bf 53.4 \\
    \cmidrule{1-8}        
        & \multicolumn{7}{c}{Base model: \texttt{llama-2-70b}}\vspace{0.25em}\\
        \iclcite{} & \bf 41.5  &	  62.9&	61.3 	&&	\bf 21.8  & 15.1 &  15.6	\\
    \cmidrule{1-8}        
        & \multicolumn{7}{c}{Base model: \texttt{ChatGPT-0301}}\vspace{0.25em}\\
        \iclcite{} &  40.4  &	 \bf  73.6&	\bf 72.5 	&&	20.8 & 20.5 & 20.9	\\
        \bottomrule
    \end{tabular}
    \caption{Comparing \agre{} on \ttsmall{llama-2-13B} against \iclcite{} on \ttsmall{llama-2-70B} and \ttsmall{ChatGPT-0301}. We directly quote results from ALCE.}
    \label{tab:compare_small_large}
\end{table}

Table~\ref{tab:compare_small_large} compares \agre{} using \ttsmall{llama-2-13B} as the base model against \iclcite{} on more capable models. We directly use the results from ALCE~\cite{alce}. Our framework is able to substantially shorten the gap between a small llama-2 model and much more capable LLMs.

\section{License of Datasets}
The licenses datasets used in our work include:
\begin{itemize}
    \item NQ~\cite{natq} under Creative Commons Share-Alike 3.0 license.
    \item StrategyQA~\cite{strategyqa} under MIT License.
    \item Fever~\cite{fever} under Creative Commons Share-Alike license.
    \item Ambiguous QA~\cite{asqa} under Creative Commons Share-Alike 3.0 license.
    \item Qampari~\cite{qampari} under Creative Commons Zero v1.0 Universal license.
\end{itemize}

\section{Additional Examples of Tuning Data}
\label{app:more_exs}
Please see Fig.~\ref{fig:app_exs1} and Fig.~\ref{fig:app_exs2} for concrete examples of tuning data.

\begin{figure}[h]
    \centering
    \footnotesize
    \begin{tabularx}{\linewidth}{|X|}
         \toprule
         \multicolumn{1}{|c|}{\bf Input} \\
        \midrule
         Task: You will be given a question and some search results. You are required to perform the following steps.\\
         \\
         First, please answer the question in 3-5 sentences, and make sure you mention relevant details in the search results. You may use the same words as the search results when appropriate. Note that some of the search results may not be relevant, so you are not required to use all the search results, but only relevant ones.  If you use the provided search results in your answer, add $[$n$]$-style citations. \\
        \\
        Next, review your response and find the unsupported sentences that do not have citations.\\
\\
Question: \\
what kind of sentence contains an independent clause and a dependent clause? \\ \\

Search Results: \\
$[$1$]$ Dependent clause\\
... A "complex sentence" contains an independent clause and at least one dependent clause. A sentence with two or more independent clauses plus (one or more) dependent clauses is referred to as a "compound-complex sentence". (Every clause contains a subject and predicate.)...\\ \\
$[$2$]$ Sentence clause structure\\
... A complex sentence has one or more dependent clauses (also called subordinate clauses). Since a dependent clause cannot stand on its own as a sentence, complex sentences must also have at least one independent clause...\\ \\
$[$3$]$ Sentence clause structure\\
A set of words with no independent clause may be an "incomplete sentence", also called a "sentence fragment". A sentence consisting of at least one dependent clause and at least two independent clauses may be called a "complex-compound sentence" or "compound-complex sentence". Sentence 1 is an example of a simple sentence...\\ \\
$[$4$]$ Topic sentence\\
Using a complex sentence is a great way to refer to the content of the paragraph above (dependent clause) and then bring in the content of the new paragraph (the independent clause). Here is a typical example: The beginning, dependent, clause probably refers to the content of a preceding paragraph that presented the ant as a community-focused worker...\\ \\
$[$5$]$ Sentence clause structure \\
This simple sentence has one independent clause which contains one subject, "I", and one verb, "run". This simple sentence has one independent clause which contains one subject, "girl", and one predicate, "ran into her bedroom"... \\
\midrule
          \multicolumn{1}{|c|}{Output} \\
          \midrule
          Answer:\\
          A complex sentence contains an independent clause and at least one dependent clause $[$1$]$. A dependent clause cannot stand on its own as a sentence, so complex sentences must also have at least one independent clause $[$2$]$. \\
          \\
          Sentences Not Supported by Citations: \\
          None. \\
         \bottomrule
    \end{tabularx}
    \caption{An example of tuning data created from NQ.}
    \label{fig:app_exs1}
\end{figure}


\begin{figure}[h]
    \centering
    \footnotesize
    \begin{tabularx}{\linewidth}{|X|}
         \toprule
         \multicolumn{1}{|c|}{\bf Input} \\
        \midrule
         Task: You will be given a question and some search results. You are required to perform the following steps.\\
         \\
         First, please answer the question in 3-5 sentences, and make sure you mention relevant details in the search results. You may use the same words as the search results when appropriate. Note that some of the search results may not be relevant, so you are not required to use all the search results, but only relevant ones.  If you use the provided search results in your answer, add $[$n$]$-style citations. \\
        \\
        Next, review your response and find the unsupported sentences that do not have citations.\\
\\
Question: \\
Could Robert Wadlow hypothetically see Frankenstein's monster's bald spot from above? \\ \\

Search Results: \\
$[$1$]$ Frankenstein's monster\\
In this version, Frankenstein gives the monster the brain of his mentor, Doctor Waldman, while his body is made from a man who killed Waldman while resisting a vaccination...In the 2004 film "Van Helsing ," the monster is shown in a modernized version of the Karloff design. He is tall, has a square bald head, gruesome scars, and pale green skin...\\ \\
$[$2$]$ Rorschach (character)\\
Rorschach is 5'6"tall and weighs 140 pounds, and, as Walter Kovacs (his"disguise"), he appears as a red-haired, expressionless, man who always carries with him a sign that reads "THE END IS NIGH"...\\ \\
$[$3$]$ Glenn Strange\\
During the wave of monster-related merchandising in the late 1950s and 1960s, Glenn Strange's iconic image often was used for the monster on toys, games, and paraphernalia, most often from his appearance in the Abbott and Costello film...\\ \\
$[$4$]$ John Bloom (actor)\\
John Bloom was an American actor. Career. Standing at 7' 4"(224 cm) tall, he is most famous for his roles as the Frankenstein Monster in the low-budget horror movie" Dracula vs. Frankenstein "(he remains the tallest actor to ever portray the character)"s...\\ \\
$[$5$]$ Model figure \\
There were also cheap plastic models for the popular market of movie monsters, comic book heroes, and movie and television characters in 1:8 size (about 9 inches or 23 cm in height)... \\
\midrule
          \multicolumn{1}{|c|}{Output} \\
          \midrule
          Answer:\\
          Yes, Robert Wadlow could hypothetically see Frankenstein's monster's bald spot from above. Frankenstein's monster is described as having a square bald head $[$1$]$. Robert Wadlow was 8 feet 11 inches tall. Frankenstein's monster was portrayed by John Bloom, who was 7 feet 4 inches tall $[$4$]$.
 \\
          \\
          Sentences Not Supported by Citations: \\
          Robert Wadlow was 8 feet 11 inches tall. \\
         \bottomrule
    \end{tabularx}
    \caption{An example of tuning data created from StrategyQA.}
    \label{fig:app_exs2}
\end{figure}

\end{document}